\documentclass[runningheads]{llncs}
\usepackage{graphicx}

\usepackage{tikz}
\usepackage{comment}
\usepackage{amsmath,amssymb} 
\usepackage{color}

\usepackage[accsupp]{axessibility}  

\usepackage[numbers,sort&compress]{natbib}

\usepackage[hidelinks]{hyperref}
\usepackage{cleveref}

\graphicspath{{./figures/}}
\DeclareGraphicsExtensions{.pdf,.jpeg,.png}
\usepackage{tikz}
\usetikzlibrary{shapes,arrows,positioning}
\tikzstyle{line}=[draw]
\usepackage[export]{adjustbox}
\usepackage[caption=false]{subfig}

\usepackage{orcidlink}

\begin{document}
\pagestyle{headings}
\mainmatter
\def\ECCVSubNumber{61}  

\title{Self-Supervised Pretraining\\ for 2D Medical Image Segmentation} 

\titlerunning{Self-Supervised Pretraining for 2D Medical Image Segmentation}
%
\author{András Kalapos\inst{1}\orcidlink{0000-0002-9018-1372} \and
Bálint Gyires-Tóth\inst{1}\orcidlink{0000-0003-1059-9822}\index{Gyires-Tóth, Bálint}}
\authorrunning{A. Kalapos and B. Gyires-Tóth}
%
\institute{Department of Telecommunications and Media Informatics\\ 
Faculty of Electrical Engineering and Informatics\\
Budapest University of Technology and Economics\\
Műegyetem rkp. 3., H-1111 Budapest, Hungary
\email{\{kalapos.andras,toth.b\}@tmit.bme.hu}}
\maketitle

\begin{abstract}
Supervised machine learning provides state-of-the-art solutions to a wide range of computer vision problems. However, the need for copious labelled training data limits the capabilities of these algorithms in scenarios where such input is scarce or expensive.
Self-supervised learning offers a way to lower the need for manually annotated data by pretraining models for a specific domain on unlabelled data. In this approach, labelled data are solely required to fine-tune models for downstream tasks. 
Medical image segmentation is a field where labelling data requires expert knowledge and collecting large labelled datasets is challenging; therefore, self-supervised learning algorithms promise substantial improvements in this field.
Despite this, self-supervised learning algorithms are used rarely to pretrain medical image segmentation networks. 
In this paper, we elaborate and analyse the effectiveness of supervised and self-supervised pretraining approaches on downstream medical image segmentation, focusing on convergence and data efficiency. 
We find that self-supervised pretraining on natural images and target-domain-specific images leads to the fastest and most stable downstream convergence. In our experiments on the ACDC cardiac segmentation dataset, this pretraining approach achieves 4-5 times faster fine-tuning convergence compared to an ImageNet pretrained model. We also show that this approach requires less than five epochs of pretraining on domain-specific data to achieve such improvement in the downstream convergence time. 
Finally, we find that, in low-data scenarios, supervised ImageNet pretraining achieves the best accuracy, requiring less than 100 annotated samples to realise close to minimal error.



\end{abstract}

\section{Introduction}
The success and popularity of machine learning in the last decade can be traced largely to the progress in supervised learning research. However, these methods require copious amounts of annotated training data. Therefore, their capabilities are limited in scenarios where annotated data is scarce or expensive. One such domain is biomedical data analysis where labelling often requires expert knowledge. In most cases, a limited number of medical professionals are available to perform certain annotation tasks and their availability for such activities is a strong constraint. Even collecting consistent and comparable unlabelled datasets at large scale is challenging due to different data acquisition equipment and practices at different medical centres. For rare diseases, the challenges of data acquisition are even greater.
The amount of available data in the medical imaging field is increasing in public datasets or national data banks (e.g., UK Biobank~\cite{UKbiobank}), however, even in these data sources, more unlabelled data could be available than with annotations that fit certain research interests. 

Self-supervised learning~\cite{JigsawSSL,ColorizationSSL,PatchSSL,DenoisingSSL,CPCv2,SimCLR,MoCo,SwAV,BYOL,BarlowTwins} (SSL) is an approach to unsupervised representation learning that attracted great research interest in recent years. Its goal is to use unlabelled data to learn representations that can be fine-tuned for a wide variety of downstream supervised learning tasks. The learned representation should be general, task agnostic and high-level to be useful for many different downstream tasks. Being able to pretrain a model on unlabelled data could significantly reduce the need for annotated data and opens the possibility to train larger and potentially more accurate models than what is feasible with purely supervised learning. 

\begin{figure}[t]
	\centering
	\includegraphics[width=0.9\linewidth]{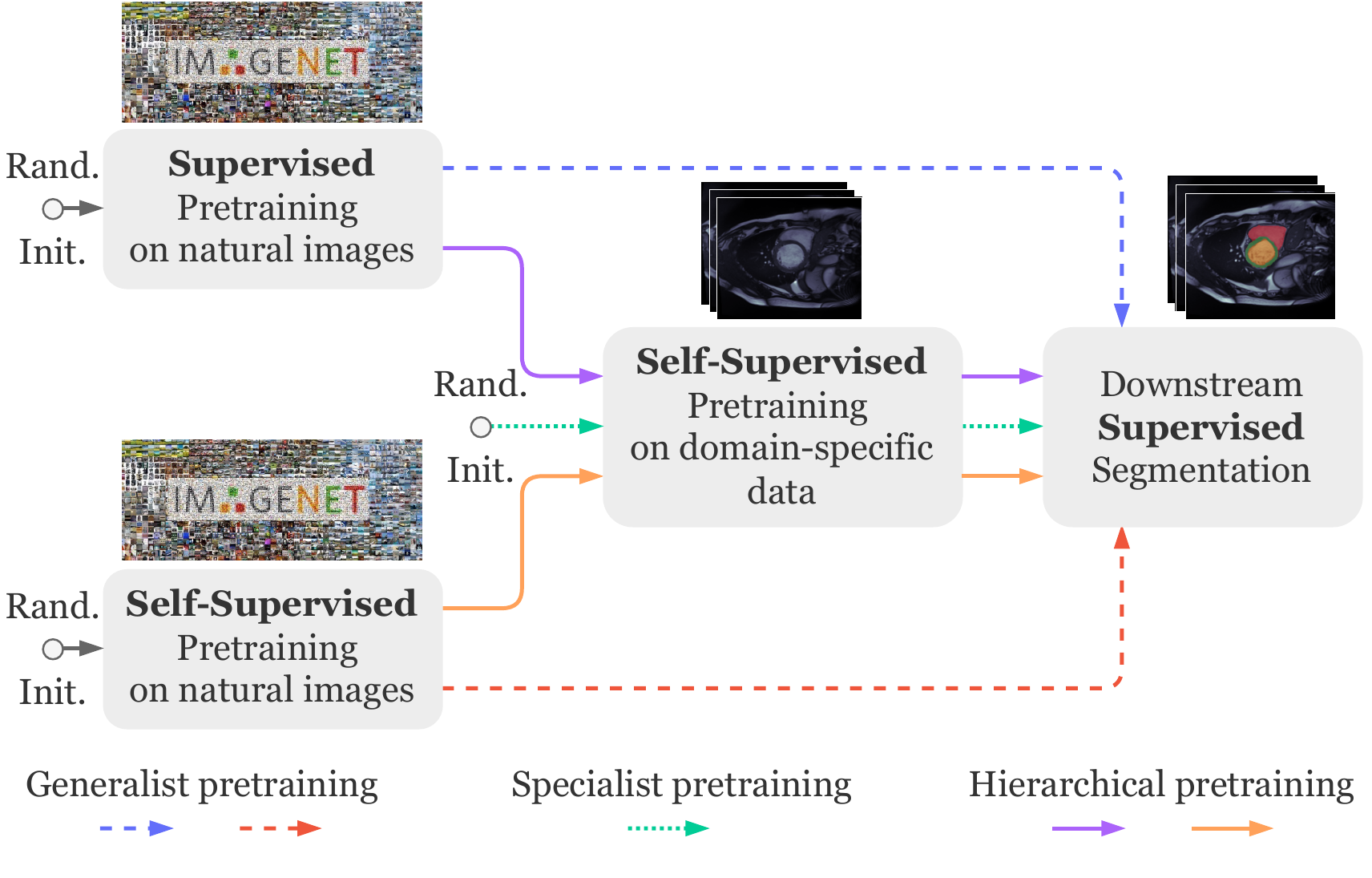}
	\caption{Pretraining pipelines that we investigate in this paper.}
	\label{fig:pretraininApproaches}
\end{figure}

In this paper, we elaborate on different supervised and self-supervised pretraining approaches and analyse their transfer learning capability on a downstream medical image segmentation task.
Investigated pretraining approaches include training steps on natural images and target-task-specific medical images. Our primary research interest is to assess the ability of self-supervised pretraining algorithms to learn representations that improve the data efficiency or accuracy on the target segmentation task. We study how downstream accuracy scales with the available labelled data in the case of different pretraining methods. The main goal of this research is to provide a guideline on the annotated training data-need of medical segmentation (focusing on cardiac imaging) with the different training approaches. The source code or our experiments is publicly available.\footnote{\url{https://github.com/kaland313/SSL-MedSeg}}  


\section{Related Work}
\subsection{Self-supervised learning}\label{sec:ssl}


Self-supervised learning methods learn representations by solving a pretext task where the supervisory signal emerges from the raw, unlabelled data. This paradigm enabled natural language processing (NLP) models with billions of parameters to learn from huge datasets and became the state-of-the-art approach for many problems in NPL\cite{GPT,BERT}. Applying transfer learning to models that were pretrained on Imagenet\cite{ImageNet} with supervised learning has been an effective and widely used method in computer vision. However, models pretrained with self-supervision are achieving state of the art for vision tasks as well, and the possibility of learning useful visual representations from uncurated and unlabelled data has been demonstrated\cite{SEER}.

However, constructing good pretext tasks for image data is not as intuitive as for text due to the high dimensionality and continuous nature of images. Early approaches used heuristic pretext tasks such as solving a jigsaw puzzle~\cite{JigsawSSL}, predicting the relative position of image patches~\cite{PatchSSL}, colourization~\cite{ColorizationSSL} and denoising~\cite{DenoisingSSL}. Recent methods (\cite{CPCv2,SimCLR,MoCo,SwAV,BYOL,BarlowTwins} etc.) rely on the instance discrimination task, which treats augmented versions of every instance from a dataset as a separate class. A set of random augmentations are applied to each image, resulting in various \textit{views} of it. The learning objective is then to assign similar latent vectors to different views of the same image, but different ones to views of different images. Methods that use such an instance discrimination task primarily differ in the set of augmentations and the approach to enforce this similarity of the latent vectors.  

These methods aim to train a model that acts as an encoder, mapping images to an embedded vector space, also called latent vectors. These encoders could then be incorporated into downstream tasks as pretrained models. 


The most common \textbf{transfer learning} approach in computer vision is fine-tuning a model that was pretrained for supervised image classification on ImageNet~\cite{ImageNet}. Another approach receiving notable attention recently is self-supervised pretraining either on ImageNet or on domain-specific data (i.e. on data that is similar or related to the problem tackled by transfer learning). Reed et~al.~\cite{HierarchicalPretraining} investigate self-supervised pretraining pipelines including ImageNet pretraining (generalist pretraining), domain-specific (or specialist) pretraining and the sequential combination of the previous two, which they refer to as "hierarchical pretraining". 

\subsection{Self-supervised learning for medical image processing}

Medical image recognition problems are commonly solved using transfer learning from natural images, even though the distribution, frequency pattern, and important features of medical images are not the same as those of natural images~\cite{Morid2021}.
Self-supervised pretraining on domain-specific data offers an alternative approach to this task, eliminating the mismatch in source and target dataset characteristics, and allowing a methodologically more precise pretraining. 
Moreover, domain-specific pretraining has the potential to reduce the number of labelled data, which has high costs in the medical domain. 

Recently, a handful of papers have been published that adapt self-supervised learning methods to pretrain models on medical image analysis problems such as histopathology~\cite{ciga2021self}, chest X-ray and dermatology classification~\cite{sowrirajan2021moco,BigSelfSupervisedModels}. These works show that MoCo~\cite{MoCo} and SimCLR \cite{SimCLR} are effective pretraining methods on medical images without major modifications to the algorithms. 
They highlight the capability of these algorithms to outperform supervised ImageNet pretraining by 1-5\% classification accuracy on various medical image recognition datasets.


\section{Methods and Data}
\subsection{Pretraining approaches}

Training pipelines consist of one or two pretraining stages followed by the downstream supervised segmentation training (see \cref{fig:pretraininApproaches}). 
All reported metrics and learning curves are from the final segmentation training step.
Between each training stage, we transfer the weights of the encoder network to the next stage, but every stage uses additional layers, which we initialize randomly. As explained in \cref{sec:sslAlgo}, the applied self-supervised learning algorithm trains an encoder through two subnetworks (called the projector and predictor). The downstream segmentation network also has an additional module after the encoder: a decoder. All these subnetworks are randomly initialized at the beginning of each training phase using Kaiming-uniform initialization~\cite{KaimingHeInit}, while the weights of the encoder are always loaded from the previous stage. In the first stage of every pipeline, the encoder is random-initialised as well.  

\subsubsection{Generalist pretraining}

In this paper, we refer to supervised and self-supervised pretraining on natural images as generalist pretraining (see \cref{fig:pretraininApproaches}). For both learning modes, this approach assumes that the representations learned on natural images are general and useful for other problems and domains. A benefit of transfer learning from natural images is the ample availability of common models pretrained on ImageNet. In our experiments, we use such pretrained models instead of running supervised and self-supervised training on ImageNet. 

\subsubsection{Specialist pretraining}
Opposing the previous assumption, we conduct experiments on self-supervised pretraining on images that come from the same domain and dataset as the labelled data for the target segmentation problem. 

\subsubsection{Hierarchical pretraining}
Hierarchical pretraining combines the previous two approaches, aiming to keep general representations learned on natural images but also fine-tune these to better match the target domain. In practice, this includes fully training a model on natural images, then continue its training on images from the target domain. 
We extend the notion of hierarchical pretraining as opposed to that of Reed~et~al.~\cite{HierarchicalPretraining}, who only refer to self-supervised pretraining on natural images as the first step of hierarchical pretraining. We elaborate on both supervised and self-supervised learning for the training step on natural images.



\subsection{Applied Self-Supervised learning method}\label{sec:sslAlgo}
\tikzstyle{encoder} = [trapezium, draw, fill=yellow!10, draw=gray!50, text width=4.5em, text badly centered, node distance=3cm, inner sep=0pt, shape border rotate=270, trapezium angle=75, rounded corners]
\tikzstyle{encoder-freezed} = [trapezium, draw, fill=gray!10, draw=gray!50, text width=4.5em, text badly centered, node distance=3cm, inner sep=0pt, shape border rotate=270, trapezium angle=75, rounded corners]

\tikzstyle{block} = [rectangle, draw, fill=yellow!10, draw=gray!50, text width=3em, text centered, rounded corners, minimum height=1.2cm]

\tikzstyle{line} = [draw, -latex', ultra thick]
\tikzstyle{line1} = [draw, -latex', ultra thick, text=orange, color=orange]
\tikzstyle{line2} = [draw, -latex', ultra thick, text=teal, color=teal]

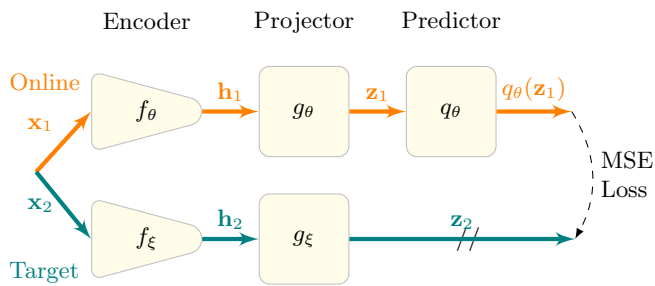
\begin{figure}[!t]
\centering
\begin{adjustbox}{max width=\linewidth}
	\begin{tikzpicture}[auto]
		\node [encoder] 				 (encoder1) {$f_\theta$};
		\coordinate[below left = 0.3333cm and 1cm of encoder1] (start0);
		\node [block, right=0.75 cm of encoder1] (mlp1)  	{$g_\theta$};
		\node [block, right=0.75 cm of mlp1] (predictor)  	{$q_\theta$};
		
		\node [encoder, below= 1 cm of encoder1] (encoder2) {$f_\xi$};
		\node [block, right=0.75 cm of encoder2] (mlp2)  	{$g_\xi$};
		
		\coordinate[right=1cm of predictor.east] (end1);
		\coordinate[right=3cm of mlp2.east] (end2);
		
		\node[draw=white, above=0.75cm of encoder1, anchor=base] {Encoder};
		\node[draw=white, above=0.5cm of mlp1, anchor=base] {Projector};
		\node[draw=white, above=0.5cm of predictor, anchor=base] {Predictor};
		\node[draw=white, above left = -0.3cm and 0.3cm of encoder1, text=orange] {Online};
		\node[draw=white, below left = -0.3cm and 0.3cm of encoder2, text=teal] {Target};
		\node[draw=white, right=1.3cm of mlp2.east] {//};
		
		\path [line1] (start0) -- node {$\mathbf{x}_1$}(encoder1.bottom side);
		\path [line1] (encoder1.top side) -- node {$\mathbf{h}_1$}(mlp1);
		\path [line1] (mlp1) -- node {$\mathbf{z}_1$} (predictor);
		\path [line1] (predictor) -- node {$q_\theta(\mathbf{z}_1)$} (end1);
		
		\path [line2] (start0) -- node [left] {$\mathbf{x}_2$}(encoder2.bottom side);
		\path [line2] (encoder2.top side) -- node {$\mathbf{h}_2$}(mlp2);
		\path [line2] (mlp2) -- node {$\mathbf{z}_2$} (end2);
		
		\path[dashed, -latex]
		(end1) edge[bend left] node [right, text width=3em] {MSE Loss} (end2);
		
	\end{tikzpicture}
\end{adjustbox}
	\caption{Architecture of BYOL for a pair of views: $\mathbf{x}_1, \mathbf{x}_2$. Modified from \cite{BYOL}}
	\label{fig:BYOLexpl}
\end{figure}
Grill et~al.~\cite{BYOL} proposed the Bootstrap Your Own Latent (BYOL) self-supervised learning method, which relies on the instance-discrimination task described in \cref{sec:ssl} and a special asymmetric architecture to avoid representation collapse. BYOL trains an encoder ($f_\theta$ on \cref{fig:BYOLexpl}) to learn the representations of another fixed encoder ($f_\xi$ on \cref{fig:BYOLexpl}) for $x_1, x_2$ different views of the same input image. In other words, the $f_\xi$ fixed network produces some $h_2$ representation of the $x_2$ augmented version of the image, which the trained network should reproduce, even though its input is a different $x_1$ augmentation of the image. Empirical evidence shows, that the trained $f_\theta$ "online" network can learn better representations than that of the fixed $f_\xi$ "target" network. This allows for an iterative update procedure where the fixed network's weights are also slowly updated, based on the improved weights of the online network. The latter one is trained by gradient-based learning, while the target network is updated using the exponential moving average of the online network’s weights. Gradients are not propagated through the target network, however, as training converges the weights of the two networks become similar and can be used for downstream tasks. 

To achieve stable training, BYOL requires a $q_\theta$ predictor network in the online branch of the network. BYOL trains the online branch using the mean squared error between the $q_\theta(\mathbf{z}_1)$ output of the predictor and the  $\mathbf{z}_2$ output of the $g_\xi$ target projector. See \cref{fig:BYOLexpl} for an illustration of BYOL's asymmetric architecture. 

\subsection{Cardiac Segmentation dataset}

The “Automated Cardiac Diagnosis Challenge” dataset~\cite{ACDCdataset} is a publicly available dataset aiming to improve deep learning-based automated cardiac diagnosis. The dataset contains cardiac MRI records from 100 patients, distributed evenly in five groups, one healthy and four with different cardiac conditions. For each patient, the complete cardiac cycle\footnote{One cardiac cycle is the period between two heartbeats.} is covered by the recordings, where each recording consists of 3D MRI scans, which include 6-18 high-resolution 2D slices. This gives approximately 25 thousand slices in total for the dataset. The resolution of the slices varies from $150\times150$ to $500\times500$ pixels. 

Segmentation labels are provided for the end-diastolic (ED) and end-systolic (ES) frames of the cardiac cycle, which gives 1900 slices with segmentation masks. For the rest of the frames, accurate segmentation labels are not available, however, self-supervised learning algorithms can leverage this data as well. We refer to the slices of annotated frames as the labelled 2D segmentation dataset and include all 25k slices (labelled and unlabelled) in the unlabelled dataset. One can consider this a semi-supervised learning dataset, with labels only available for a subset of all slices.

\begin{figure}[!t]
	\centering
	\includegraphics[width=0.8\linewidth]{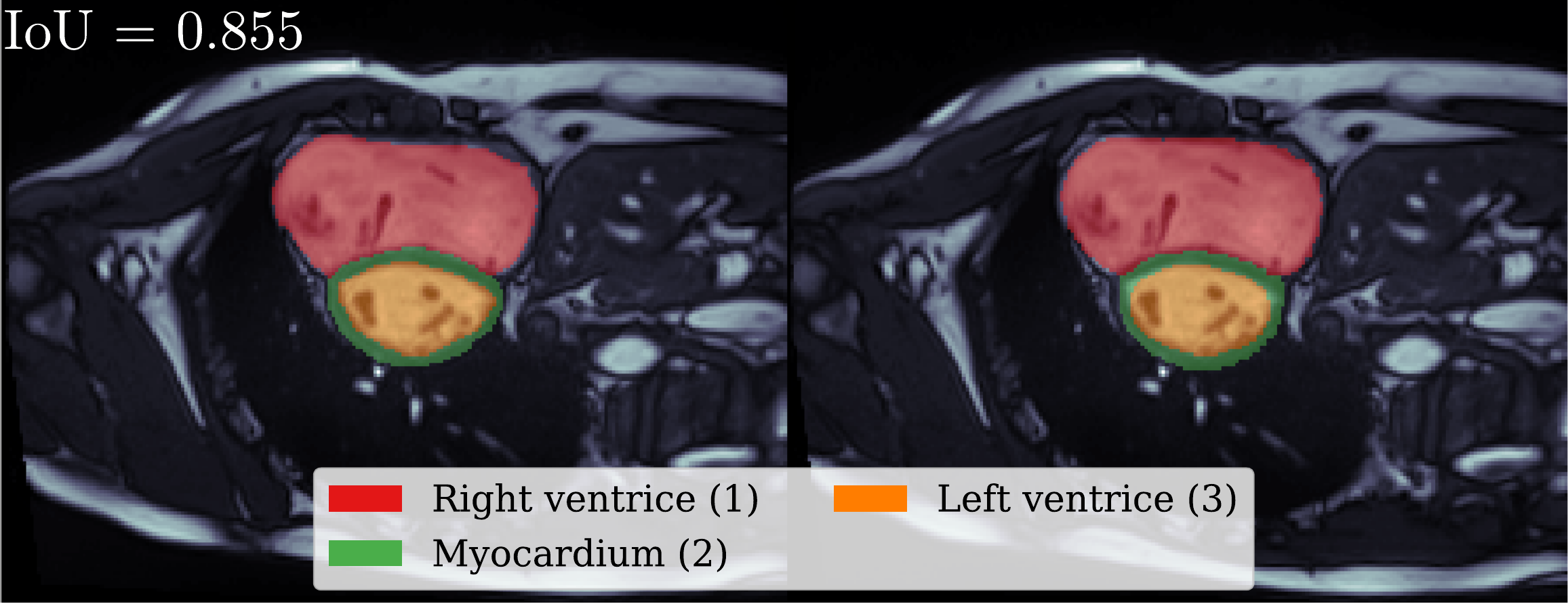}
 
	\caption{Left: An example image from the ACDC dataset\cite{ACDCdataset} with ground truth segmentation masks overlaid. Right: The same input image with predictions of a model that was pretrained using specialist pretraining (BYOL\cite{BYOL}).}
	\label{fig:acdcExample}
\end{figure}

Segmentation masks are provided with four classes, left and right ventricle (LV, RV), myocardium (MYO), and background (see \cref{fig:acdcExample} as an example).

\subsection{Model architecture}\label{sec:modelArchitecture}

We selected a U-Net\cite{UNet} architecture for the segmentation training due to the successful application of this architecture for medical image segmentation. The architecture and implementation of the encoder match with the ones we pretrain, which is necessary for transferring the pretrained weights to the segmentation model. Encoders pretrained on ImageNet expect three-channel, RGB images as input, however, MRI scans are monochromatic. Therefore, the first layer of these encoders needs to be adjusted for single-channel input, which we achieve by depth-wise summation of the weights of the first layer (after~\cite{PyTorchImageModels}). 

\section{Experiments}

Our primary research interest is to assess the scaling of downstream accuracy with the amount of available labelled training data for different pretraining strategies. We hypothesize that efficient pretraining approaches improve the accuracy of downstream training for limited data.

\subsection{Training phases}

\subsubsection{Natural images}
In all our experiments, we pretrained ResNet-50 \cite{ResNet} encoders, which we also used as the encoder of a U-Net segmentation network. We selected this architecture for its frequent use in self-supervised learning literature. 
We used ~\cite{PyTorchImageModels} instead of training an encoder from scratch in our experiments, which utilized supervised ImageNet pretraining.
For BYOL pretraining on ImageNet, we obtained weights from a third-party GitHub repository\footnote{\url{https://github.com/yaox12/BYOL-PyTorch}; Accessed on 4 May 2022}. This model was trained for 300 epochs on 32 NVIDIA V100 GPUs. 

\subsubsection{Domain-specific}
For self-supervised pretraining on the ACDC dataset, we apply simple augmentations that are reasonable on monochromatic medical images: random resizing and cropping, horizontal flipping and brightness, and contrast perturbations. For this training phase we adapt the augmentations to the ACDC dataset, but keep most hyperparameters as published in~\cite{SoloLearn}.

In experiments with ImageNet pretraining preceding domain-specific SSL training, we adapt the weights of the input layer from three-channel to single-channel as explained in \cref{sec:modelArchitecture}. In these experiments, we run SSL pretraining for 25 epochs, which takes 45 minutes on a single NVIDIA V100 GPU.
For specialist pretraining, we randomly initialize the encoder at the beginning of the SSL training and run it for 400 epochs (which takes almost 10 hours).  

\subsubsection{Downstream segmentation}
We apply the same augmentations as in the domain-specific SSL stage and train for 150 epochs to reach convergence in all experiments, which takes 20 minutes on average with one V100 GPU. Furthermore, we use Jaccard loss and periodic cosine annealing learning rate schedule.
The U-Net's decoder is relatively small with 9 million trainable parameters, compared to the encoder's 23 million. With such a small decoder we expect that segmentation performance relies more on the encoder than in the case of a symmetric architecture with a large encoder. The decoder uses nearest neighbour upsampling and two convolutional layers at each upsampling stage.
We implement downstream segmentation training based on an open-source repository for segmentation models and methods~\cite{SegmentationModelsPytorch}.

\subsection{Data-efficient learning}
We aim to test the hypothesis that efficient pretraining improves the downstream accuracy in low-annotated-data scenarios.
To test this hypothesis, we run segmentation training on subsets of the labelled dataset, ranging from 1 sample to the full dataset. We also vary the pretraining strategy preceding the segmentation training. 
Each training runs for the same number of steps which is equivalent to 150 epochs on the full dataset and each run was repeated with 10 different randomly set initial parameter states (i.e. random seeds).


\section{Results}

\begin{figure*}[!t]
	\centering
	\subfloat[\label{fig:resutlsLearningCurve}]{
	\includegraphics[height=4.5cm,valign=t]{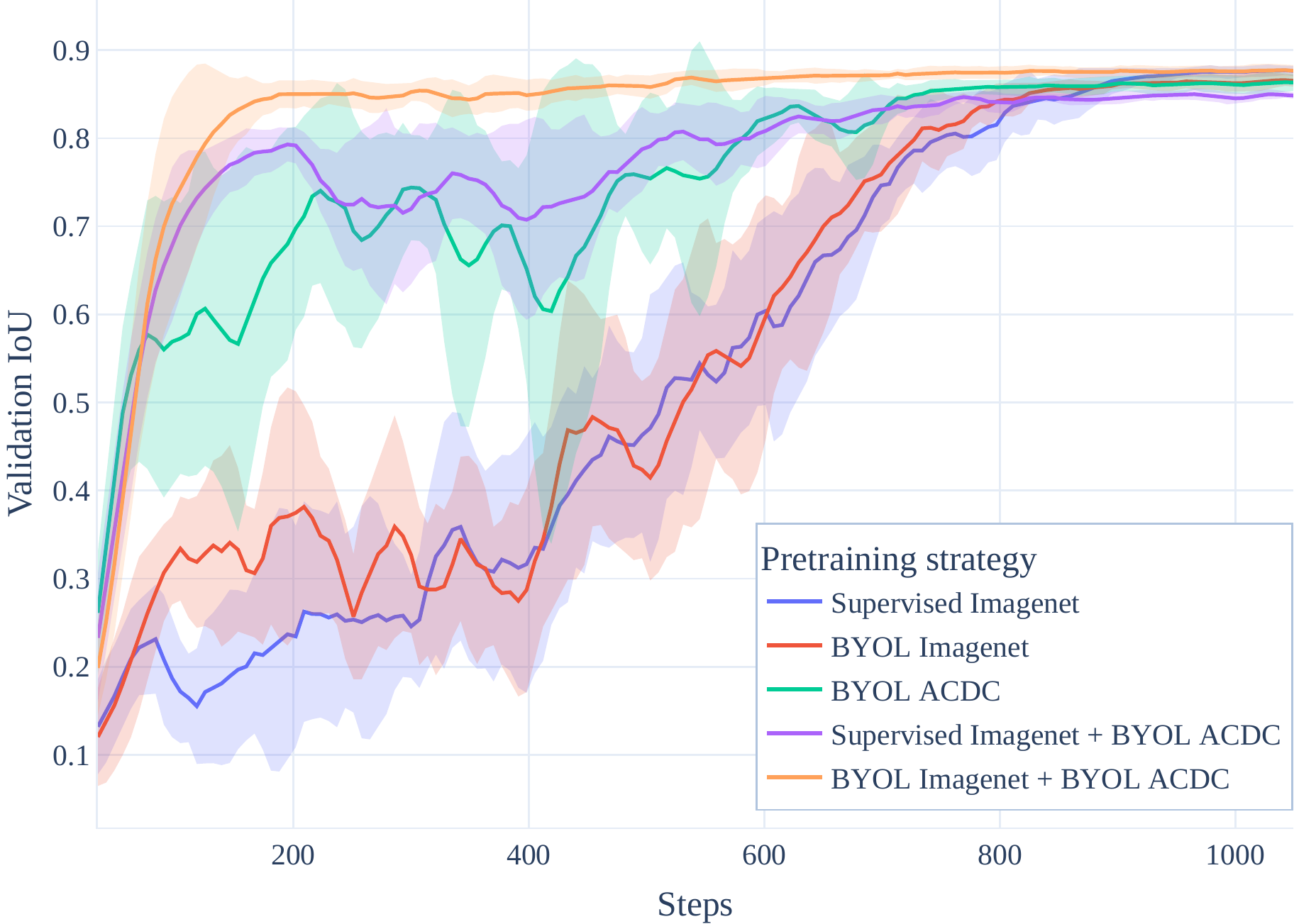}
	}
	\subfloat[\label{fig:resutlsLearningCurveAuC}]{
	\includegraphics[height=5cm,valign=t]{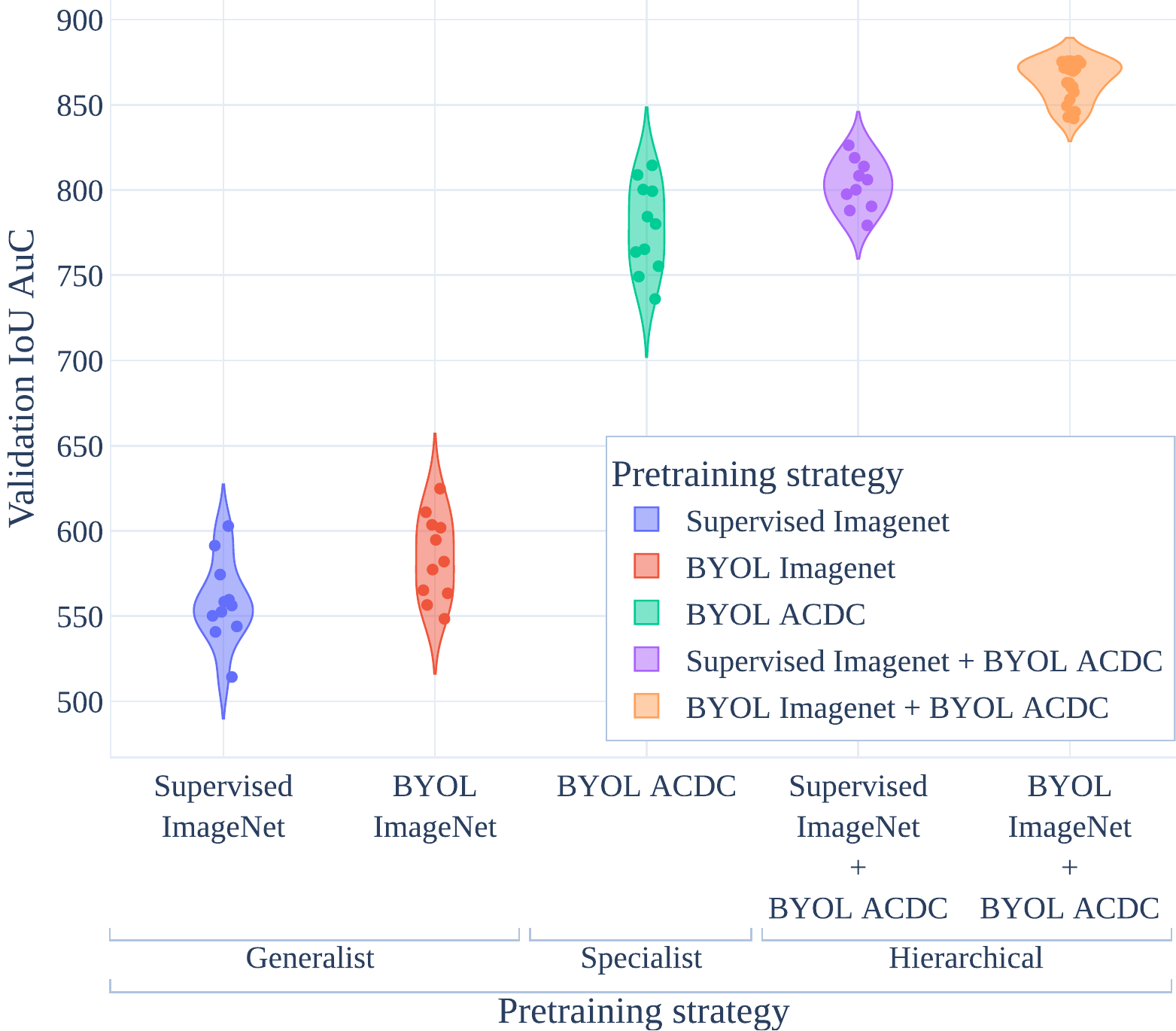}} 
	\caption{(a) Learning curves for segmentation training on the full labelled dataset, preceded by different pretraining pipelines. Solid lines show means, shaded area represents $\pm$ standard deviation of 10 runs. 
	(b) Violin plot of the area under the learning curves on the left. Ordered according to medians. Colours for different pretraining pipelines are consistent across all figures.} 
	
\end{figure*}

\begin{figure*}[!t]
	\centering
	\includegraphics[width=\textwidth]{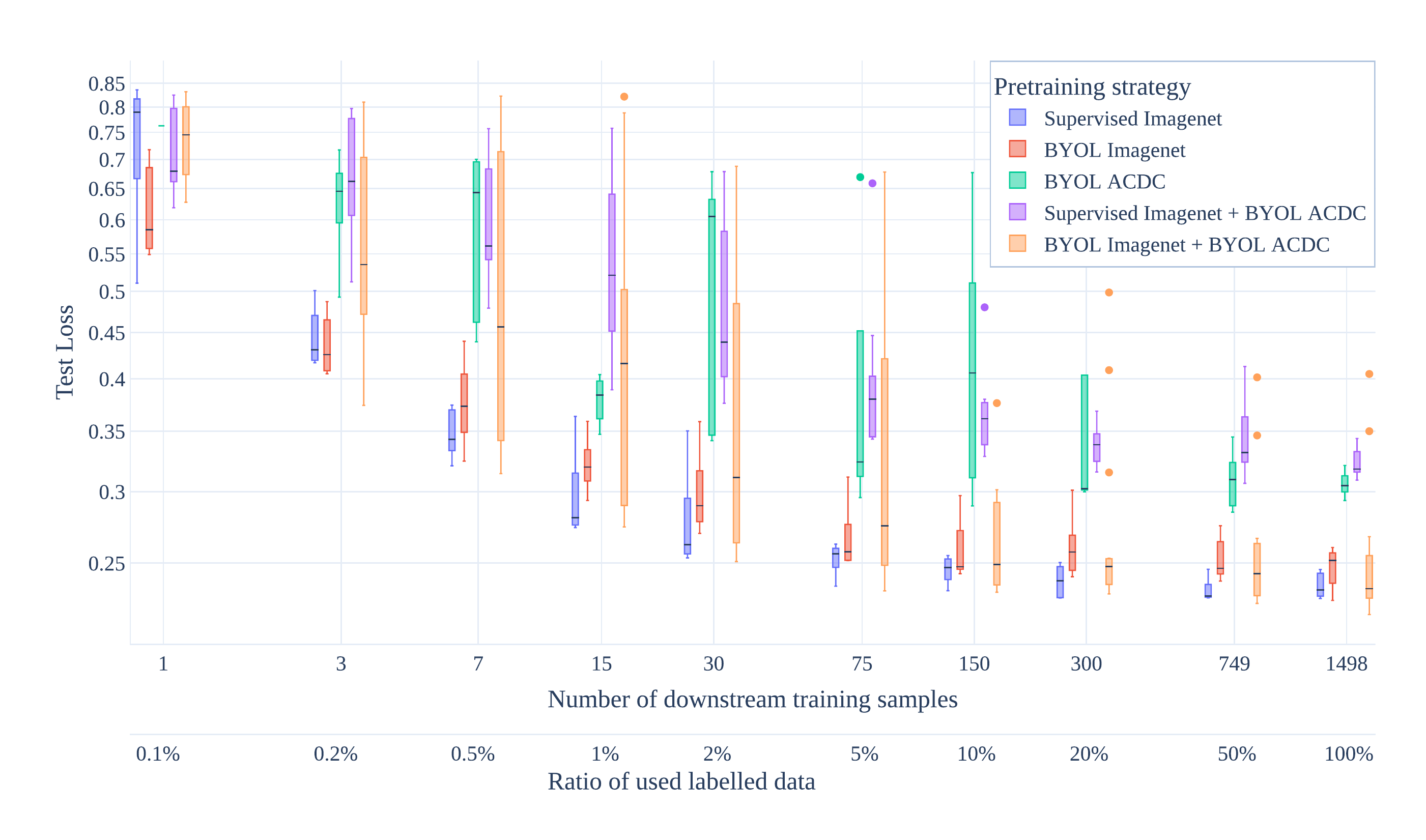}
	\caption{Downstream segmentation test loss (Jaccard loss), for different annotated data quantities and pretraining pipelines. We select training samples, which are individual MRI slices randomly from slices of all patients. Each box represents statistics of 10 runs with different random seeds. Boxes are grouped for each data setting, resulting in offsets from the true x-axis location, which must be taken into consideration when interpreting the figure. }
	\label{fig:resultsDataefficient}
\end{figure*}

\subsection{Convergence and stability}\label{sec:resultsConvergence}

\Cref{fig:resutlsLearningCurve} shows learning curves for downstream segmentation training with different pretraining approaches. The figure shows that hierarchical pretraining methods converge faster than generalist or specialist pretraining on this semantic segmentation task. This indicates that representations learned during both stages retain information useful for the target task. However, two notable differences separate these hierarchical methods as well. First, pretraining involving only self-supervised steps leads to faster and more stable convergence compared to supervised ImageNet pretraining followed by domain-specific SSL pretraining. Second, the final segmentation Jaccard index (Intersection over Union, IoU score) for the latter approach is significantly lower than for any other method. \Cref{fig:resutlsLearningCurve} also reveals that self-supervised only hierarchical pretraining (BYOL ImageNet + BYOL ACDC) leads to downstream convergence in $\sim$200 steps, while other methods reach close to maximal IoU in 4-5 times more steps. 
The area under the learning curves (see \cref{fig:resutlsLearningCurveAuC}) highlight the stability of self-supervised only hierarchical pretraining even more clearly (BYOL ImageNet + BYOL ACDC on the figure). This figure also indicates that specialist self-supervised pretraining (BYOL ACDC) leads to similar results as mixing supervised and self-supervised pretraining (Supervised ImageNet + BYOL ACDC). 
Our findings show that among the investigated approaches, self-supervised only pretraining (i.e. BYOL ImageNet + BYOL ACDC pretraining) produces representations that achieve the fastest and most stable downstream convergence.

\subsection{Data-efficiency}
\Cref{fig:resultsDataefficient} shows how downstream error scales with the available annotated data, in the case of different pretraining methods. Contradicting our hypothesis, the figure reveals that with low data quantities generalist pretraining approaches have lower test error compared to hierarchical pretraining. 

Hestness et~al.~\cite{PowerLawOfDL} investigates similar scaling behaviour as visible on \cref{fig:resultsDataefficient}. They show the existence of a power-law region, where the reduction of test error follows a power-law scaling with increasing data quantities. Hestness et~al.~\cite{PowerLawOfDL} also point out that by increasing the amount of training data above a certain threshold, the improvement of test error starts to decrease slower than a power-function of the data quantity. We can observe both scaling regions on \cref{fig:resultsDataefficient}. Below 2\% of the total available training data, that is ~30 MRI slices, generalist pretraining approaches seem to follow power-law scaling (linear scaling on a log-log plot), while above 5\% of available training data, the reduction of error is notably slower, indicating the transition to the irreducible error regime. The amount of training data that corresponds to this transition is important to machine-learning development and decision-making, because investing more in annotated data above this region has a much smaller return in error reduction, than with fewer data. In simple terms, labelling a given number of additional samples is increasingly less profitable above this transition region. In our experiments this transition region is around 30-75 randomly selected slices of MRI records, annotating more slices still improves downstream performance, however, the expected improvement after adding an annotated sample reduces. 

\begin{figure}[t]
	\centering
	\includegraphics[width=0.8\linewidth]{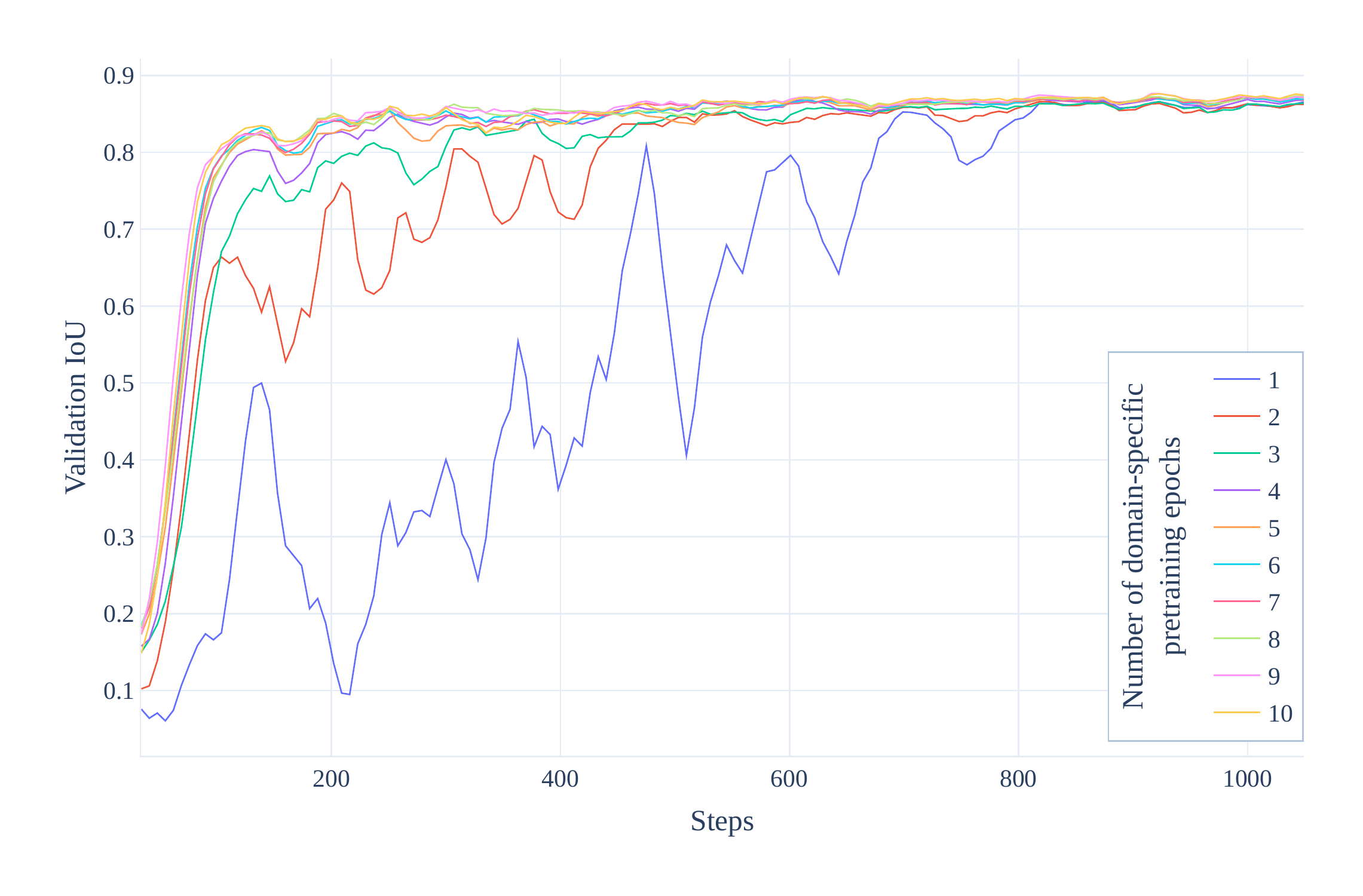}
	\caption{Downstream learning curves after different number of domain-specific pretraining epochs in a hierarchical pretraining pipeline. Starting with a model pretrained using BYOL on ImageNet, we run BYOL pretraining on the ACDC dataset, save the encoder after every epoch, and execute downstream training based on each of these. The figure shows learning curves for these downstream segmentation trainings.}
	\label{fig:pretrainingEpochs}
\end{figure}

\subsection{Domain-specific pretraining epochs}

We experiment with running the domain-specific step for a different number of epochs. As shown on \cref{fig:pretrainingEpochs}, we find that 3-4 epochs are sufficient to achieve the fast-converging, stable learning curves that we presented in \cref{sec:resultsConvergence}.

\section{Conclusion}

In this work we investigate supervised and self-supervised approaches to pretraining convolutional encoders for medical image segmentation. In addition, we study how these affect the data efficiency of the downstream training. 
We found that combining self-supervised learning on natural images and target-task-specific images (in other words hierarchical pretraining) leads to fast and stable downstream convergence, 4-5 times faster compared to the widely used ImageNet pretraining.
We also found that, with proper pretraining, only a few dozen annotated samples achieve good segmentation results on cardiac MRI segmentation. 
In our experiments on the ACDC dataset, 30-75 randomly selected annotated slices are sufficient to achieve close to minimal error.



\section*{Acknowledgment}
The research presented in this paper, carried out by Budapest University of Technology and Economics has been supported by the Hungarian National Laboratory of Artificial Intelligence funded by the NRDIO under the auspices of the Hungarian Ministry for Innovation and Technology.
We thank for the usage of the BME Joker and ELKH Cloud GPU infrastructure (\url{https://science-cloud.hu/}) that helped us achieve the results published in this paper. We gratefully acknowledge the support of NVIDIA Corporation with the donation of the NVIDIA GPU also used for this research. 

\clearpage
%
%
\bibliographystyle{splncs04}
\bibliography{Cardiac-Paper}
\end{document}